\documentclass{article}

\usepackage{amsmath}
\usepackage{amsthm}
\usepackage{mathtools}
\usepackage{xargs}
\usepackage{subfig}

\PassOptionsToPackage{numbers, compress}{natbib}
\usepackage[preprint]{neurips_2020}

\usepackage[utf8]{inputenc} 
\usepackage[T1]{fontenc}    
\usepackage{hyperref}       
\usepackage{url}            
\usepackage{booktabs}       
\usepackage{amsfonts}       
\usepackage{nicefrac}       
\usepackage{microtype}      
\usepackage{grffile}

\begin{document}

\newcommand{\bb}[1]{\mathbb{#1}}

\newcommand{\freqone}[1]{\ensuremath{f_{#1}}}
\newcommand{\freqtwo}[2]{\ensuremath{f_{#1,#2}}}
\newcommand{\freqthree}[3]{\ensuremath{f_{#1,#2,#3}}}
\newcommand{\pr}[3]{\ensuremath{\phi_{#1,#2,#3}}}
\newcommand{\prs}[2]{\ensuremath{\phi_{#1,#2}}}
\newcommand{\prva}{\ensuremath{\phi_\alpha}}
\newcommand{\prvb}{\ensuremath{\phi_\beta}}
\newcommand{\corr}[3]{\ensuremath{\Gamma_{#1,#2,#3}}}






\title{Algebraic Ground Truth Inference: Non-Parametric Estimation of Sample Errors by AI Algorithms}

%

\author{%
  Andr\'es Corrada-Emmanuel 
    \\
  Data Engines\\
  \texttt{andres.corrada@dataengines.com} \\
  \And
  Edward Pantridge\\
  Swoop \\
  \texttt{ed@swoop.com} \\
  \AND
  Edward Zahrebelski \\
  Swoop \\
  \texttt{eddie@swoop.com} \\
  \And
  Aditya Chaganti \\
  Swoop \\
  \texttt{aditya@swoop.com} \\
  \And
  Simeon Simeonov \\
  Swoop \\
  \texttt{sim@swoop.com} \\
}

\maketitle

\begin{abstract}
Binary classification is widely used in ML production systems. Monitoring classifiers 
in a constrained event space is well known. However, real world production systems often
lack the ground truth these methods require. Privacy concerns may also require that the 
ground truth needed to evaluate the classifiers cannot be made available.
In these autonomous settings, non-parametric estimators of performance are an attractive
solution. They do not require theoretical models about how the classifiers made errors in 
any given sample. They just estimate how many errors there are in a sample of an industrial 
or robotic datastream. We construct one such non-parametric estimator of the sample errors for an 
ensemble of weak binary classifiers. Our approach uses algebraic geometry to
reformulate the self-assessment problem for ensembles of binary classifiers as
an exact polynomial system. The polynomial formulation can then
be used to prove - as an algebraic geometry algorithm - that no general solution to the 
self-assessment problem is possible. However, specific solutions are possible in
settings where the engineering context puts the classifiers close to independent errors.
The practical utility of the method is illustrated on a real-world dataset from an online 
advertising campaign and  a sample of common classification benchmarks. 
The accuracy estimators in the experiments where we have ground truth are 
better than one part in a hundred. The online advertising campaign data, 
where we do not have ground truth data, is verified by an internal
consistency approach whose validity we conjecture as
an algebraic geometry theorem.
We call this approach - \emph{algebraic ground truth inference}
\end{abstract}

\section{Introduction}

The problem of self-assessment is everywhere in science and technology. Here
we consider the self-assessment problem for an ensemble of binary classifiers.
Is it possible to create a non-parametric estimator for the ensemble errors when
we do not have the ground truth for their decisions? Yes. The constructive algorithm
we present uses ideas from algebraic statistics, data streaming algorithms and
errror-correcting codes to compute an estimate for the ground truth statistics of
the errors made by the classifiers on the given sample.

It may seem strange that we could estimate, without any theory of how the classifiers
made those errors, a statistic of the errors when we do not have the ground truth
for their correctness of their decisions. It certainly runs counter to all previous
published approaches (e.g. \citet{Raykar2010}, \citet{Liu2012}, \citet{Zhou2012},
\citet{Zhang2014})
to ground truth inference (GTI) - the estimation of sample statistics
that require the ground truth.

\subsection{What is Ground Truth?}

The term \emph{ground truth} refers to the correct decisions the AI algorithm
should have made on the given sample. One could define the algebra presented here as being
generally applicable to developing self-assessment algorithms when one wants
to estimate the error of noisy random estimators of ground truth. In the case
of binary classification, the ground truth is the correct label for each output
of the noisy estimator of the ground truth - the binary classifier.

\subsection{Why estimation of ground truth statistics?}

The motivation for GTI is simple and analogous to that used in data streaming
algorithms. In commercial or scientific settings, there is frequently one
or two ground truth statistics that are valuable. For example, speech recognition
research and commercial development are mostly driven by a single statistic of
performance - Word Error Rate (WER). Similarly, when deploying binary classifiers
in an industrial AI pipeline, we can be satisfied with just knowing the average
accuracy of the classifiers. Therefore, we have observed that in many settings
the role of curated data with ground truth annotations is to carry out the correct
calculation of a sample ground truth statistic. If so, why not dispense with
getting the ground truth and just estimate the wanted statistic?

This approach is also taken by data streaming algorithms like HyperLogLog. They forgo
accumulating the ground truth for the data and instead sketch it for the purposes
of then estimating a wanted statistic. In the case of HyperLogLog, it estimates
the count of distinct values observed while refusing to remember the ground truth
for the data (the full set of distinct values so far observed).

\subsection{The advantages and limitations of non-parametric estimation of AI errors}

Taking a non-parametric approach to error estimation has many attractive features:
\begin{itemize}
  \item It bypasses the problem of understanding what model describes the way the classifiers are making errors. This means you do not need a meta-model of how the errors are being made.
  \item It has a very small memory footprint. The binary classifier self-assessment sketch for three classifiers consists of eight integer counters.
\end{itemize}

Non-parametric estimation has one large theoretical disadvantage.
No model of the data stream or the performance of the binary classifiers is obtained. Having
measured the error in a sample, we have just that - a single number. There is no model to
predict future or past errors. There is no model to potentially assign causes to errors. 
Algebraic GTI only estimates the sample error. In that way it is no different
than other non-parametric estimators like HyperLogLog or Good-Turing smoothing.

Our approach to GTI differs from the previous literature on the subject.
The concept that it is possible to use an ensemble's decisions to estimate each classifier's error
on the sample was first described by \citet{Dawid79}. The topic of GTI enjoyed a resurgence
in the 2010s with many papers \citep{Raykar2010,Wauthier2011,Liu2012,Zhou2012,Zhang2014}
presented in NeurIPS and in other venues. An extensive 2017 survey by \citet{Zheng} of these
approaches concluded that they were not stable across different domains. This is
our main critique of these methods - since they are Bayesian, they rely on carefully selected hyper-parameters to estimate the uncertainty of the model they fit. But if we don't
have ground truth, how do we know that a particular hyper-parameter setting is correctly
capturing the errors in the sample? We do not. The method presented here should be
viewed, not as substitute for Bayesian models, but as a complementary tool that
can independently assess the errors on a given sample. The estimates of algebraic
GTI can then be used to guide a more appropriate hyper-parameter setting.

Another critique of Bayesian methods for self-assessment is that they will always
return a best fit. This is to be contrasted with the algebraic approach presented
here where polynomial root solutions may be outside the unit range or imaginary - 
a clear alarm that the assumptions inherent to the approach are violated and cannot explain observed classifier decisions.

\subsection{Algebraic methods for non-parametric estimation of AI errors}

In the 1990s \citet{Pistone}
pioneered the use of algebraic geometry in statistics
by recasting traditional statistical tasks, such as experimental
design, as polynomial problems.

The technique presented here also uses the math of algebraic geometry
for a statistical task. Nevertheless, unlike previous Algebraic Statistics
work, this technique is not concerned with models about contingency table observations.
Instead, it estimates true counts. Furthermore, there is no Algebraic Statistics inside of
Algebraic GTI, which is analogous to saying that Algebraic GTI can be used for Algebraic Statistics but does not rely on Algebraic Statistics for the method to work.

\section{Algebraic Geometry Of A Single Binary Classifier}

Before tackling the full problem of an ensemble of binary classifiers,
we discuss a simpler case - the single classifier - to illustrate the
language of algebraic geometry and how it relates to the problem of
self-assessment for a single classifier.

\subsection{The polynomial system relating observable statistics
to ground truth statistics}

The black-box approach of non-parametric estimation means that the only
observable sample statistic for the classifier's label decisions are
the two counts, \freqone{\alpha} and \freqone{\beta} - the frequency of times
the two labels, $\alpha$ and $\beta$ where observed in the sample. The
sample values of these sample statistics are integer ratios. Since
these sample statistics are readily observed without needing knowledge
of the ground truth, we call them \emph{observables}.

These point statistics are not the only ones possible for a given sample.
For example, we may be concerned with sequence statistics when reconstructing
a DNA string. Thus our work here should not be considered a
general solution to any possible statistics for binary classifiers but as a solution for the self-assessment problem for point statistics. In the Conclusion, we discuss the implications of this for future research.

We write the two observable statistics for a single classifier in terms of 
the 5 unknown ground truth statistics of this ensemble describing $n=1$ as a polynomial system as follows,
\begin{align}
\freqone{\alpha}& = \prva \pr{1}{\alpha}{\alpha} + \prvb \pr{1}{\beta}{\alpha}\\
\freqone{\beta}& = \prva \pr{1}{\beta}{\alpha} + \prvb \pr{1}{\beta}{\beta}\\
1& = \freqone{\alpha} + \freqone{\beta}\\
1& = \prva + \prvb\\
1& =  \pr{1}{\alpha}{\alpha} + \pr{1}{\beta}{\alpha}\\
1& = \pr{1}{\beta}{\beta} +  \pr{1}{\alpha}{\beta}.
\end{align}
Here the \pr{1}{\ell_j}{\ell_i} variables are the unknown frequencies that classifier
1 identified $\ell_i$ as $\ell_j.$ The \prva \, and \prvb \, variables are the unknown prevalences of the true labels in the sample. Ideally, we want to solve for these statistics generally in terms of $\freqone{\alpha}$ and $\freqone{\beta}$.

All the $\phi$ sample statistics fall into the 2nd category of sample statistics
we call \emph{ground truth statistics}. These are sample statistics that require knowledge
of ground truth to compute. With ground truth, these sample statistics can be readily computed. The job of Algebraic GTI is to estimate these ground truth statistics when they are missing.

Note, furthermore, that the ground truth statistics (the $\phi$ variables) are
themselves divided into two classes: the environmental ground truth statistics
(the prevalence of the labels) and performance ground truth statistics (the
accuracy of the classifiers).

The polynomial system above can be simplified by eliminating some of the variables via
the normalization equations, which results in the following simpler polynomial
system,
\begin{align}
\freqone{\alpha}& = \prva \prs{1}{\alpha} + (1 - \prva) (1 - \prs{1}{\beta})\\
\freqone{\beta}& = \prva (1 - \prs{\alpha}) + (1 - \prva) \pr{1}{\beta}{\beta}.
\end{align}
Terms like \prs{i}{\ell} represent the accuracy of the ith classifier on the $\ell$ label.

Algebraic geometry is concerned with studying the geometric objects that correspond
to zeros of a polynomial system. The set of points in the polynomial variables space
that satisfy these polynomials is called an \emph{algebraic variety}. In the case
of one binary classifier that geometric object is a surface in the three dimensional
space of the (\prva, \prs{1}{\alpha}, \prs{1}{\beta}) variables as shown in Figure 1(a).

\begin{figure}
    \centering
    \subfloat[One classifier]{{\includegraphics[scale=.20]{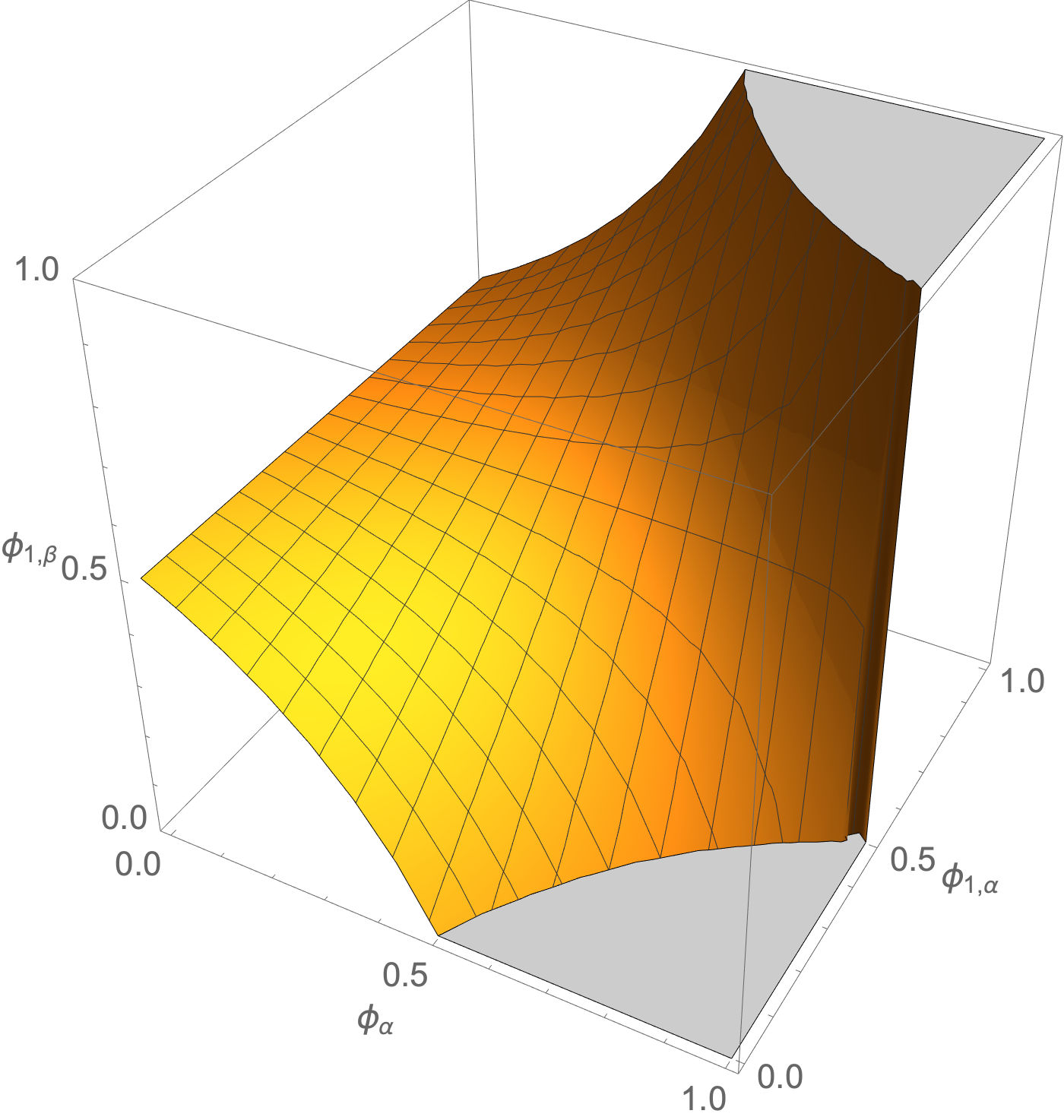} }}
    \subfloat[Two classifiers]{{\includegraphics[scale=.20]{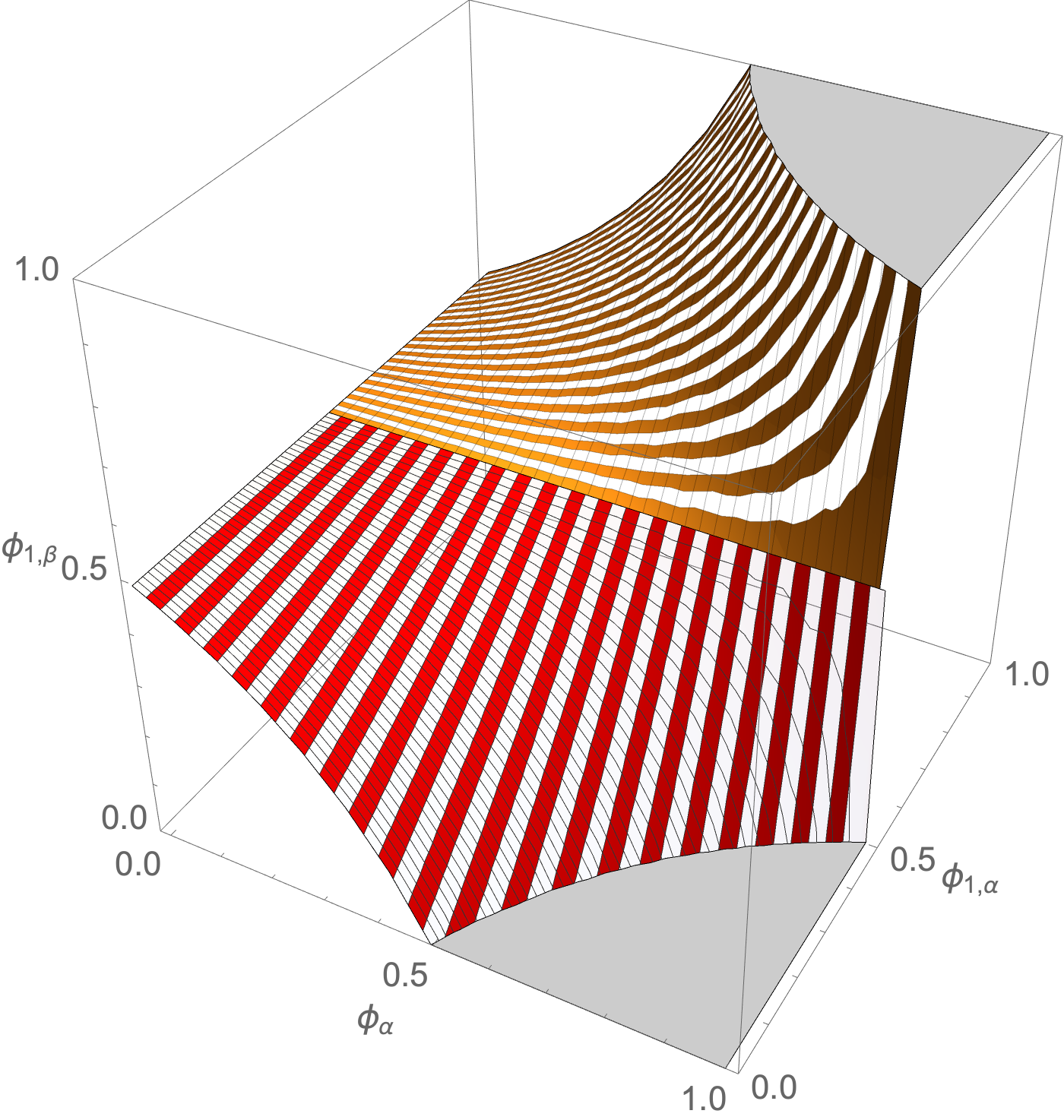} }}
    \subfloat[Three classfiers]{{\includegraphics[scale=.20]{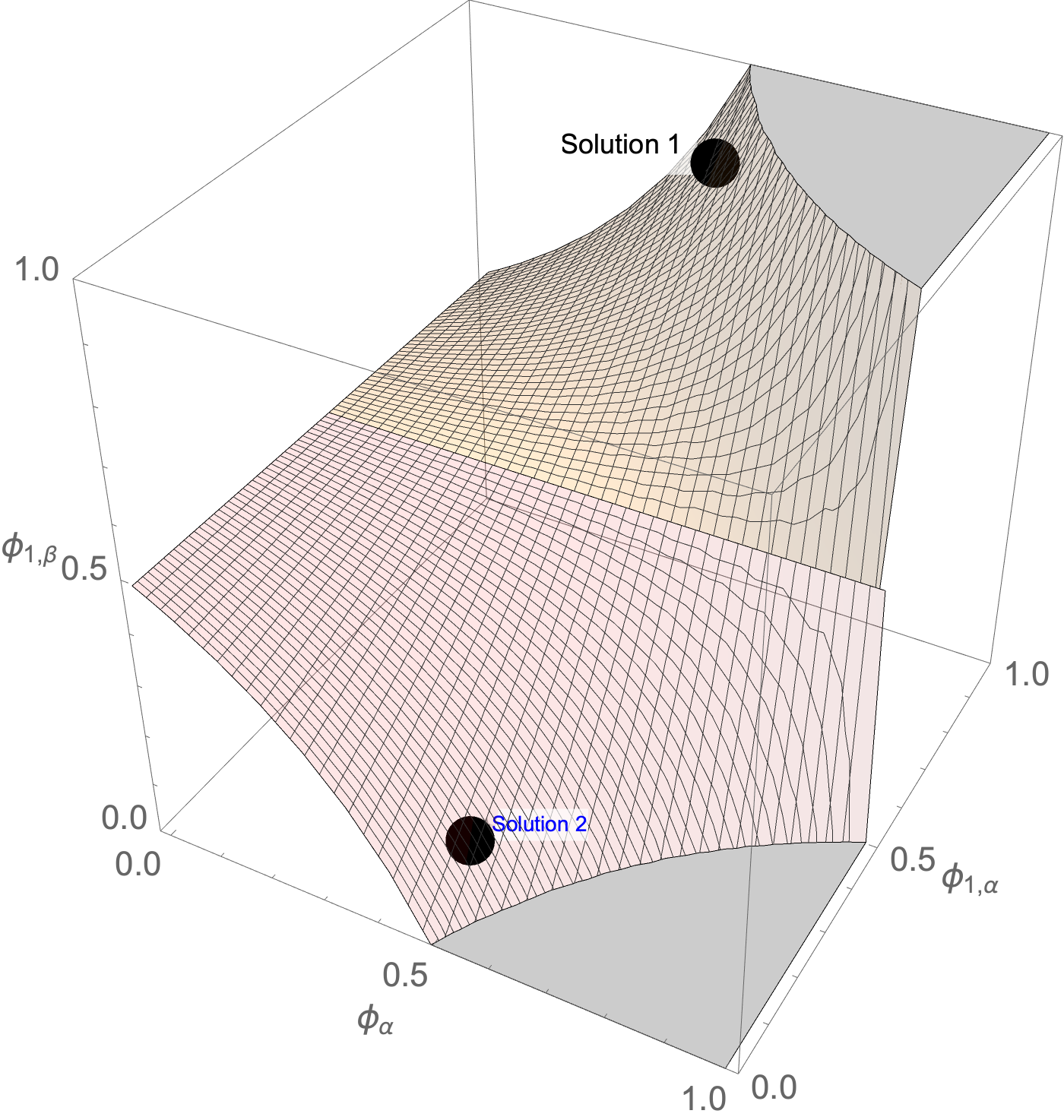} }}

    \caption{The algebraic varieties for the self-assessment polynomials of varying numbers of classifiers in the
    {\tt twonorm} experiment. As more classifiers are added in the self-assessment, the uncertainty
    in the ground truth statistics decreases to two points when three or more are used. Moving left to right, the algebraic variety transforms from a surface in the single classifier case to a fractured surface in the two classifier case and finally collapses into two individual points in the three classifier case.}%

    \label{figure:tryptich}

\end{figure}

Since the algebraic variety is a surface for one binary classifier, we conclude
that a single binary classifier cannot self-assess. The formal way to state
this is as a property of the Gr\"obner basis for equations 7 and 8
computed with the elimination order on monomials. Details are discussed in Supplement A.

\section{Algebraic Geometry of Self-Assessment for Two Binary Classifiers}

We now consider an $n=2$ ensemble of binary classifiers. Just like the
case of $n=1$, we can write an exact polynomial description
of the sample statistics. But having moved to using two binary classifiers,
we must account for non-independence between the errors made by each classifier within
a given sample.

While the formalism presented here can be extended to a classification of 3
or more labels by using indicator functions, binary classification 
allows us to think of non-independence in terms of covariance measures. 
In the case of two binary classifiers this
requires that we introduce two new ground truth statistics defined as follows,
\begin{equation}
    \corr{i}{j}{\ell} = \frac{1}{S} \sum_{d=1}^{S} 
\left(x_{d,i} - \bar{x}_i \right) \left(x_{d,j} - \bar{x}_j \right),
\end{equation}
where $S$ is the size of the sample. The $x_{d,i}$ variables are 0/1 variables
indicating whether the $i$ classifier correctly classified the item $d$ in the
sample. The $\bar{x}_i$ variable is the average performance of the classifier
on the given label and sample. These happen to be identical to the \pr{i}{\ell}{\ell}
statistics we are trying to estimate.

We can now write the full, \emph{exact} polynomial system for two binary classifiers,
\begin{align}
\freqtwo{\alpha}{\alpha} & = \prva \left( \prs{1}{\alpha}  \, \prs{2}{\alpha} + \corr{1}{2}{\alpha} \right) + \
         (1 - \prva) \left( (1 - \prs{1}{\beta}) (1 - \prs{2}{\beta}) + \corr{1}{2}{\beta} \right)\\
\freqtwo{\alpha}{\beta} & = \prva \left( \prs{1}{\alpha}  \left( 1- \prs{2}{\alpha} \right) - \corr{1}{2}{\alpha} \right) + \
         (1 - \prva) \left( (1 - \prs{1}{\beta}) \prs{2}{\beta} - \corr{1}{2}{\beta} \right)\\
\freqtwo{\beta}{\alpha} & = \prva \left( (1 - \prs{1}{\alpha})  \prs{2}{\alpha} - \corr{1}{2}{\alpha} \right) + \
         (1 - \prva) \left( \prs{1}{\beta} (1 - \prs{2}{\beta}) - \corr{1}{2}{\beta} \right)\\
\freqtwo{\beta}{\beta} & = \prva \left( (1 - \prs{1}{\alpha})  (1 -\prs{2}{\alpha}) + \corr{1}{2}{\alpha} \right) + \
         (1 - \prva) \left( \prs{1}{\beta} \, \prs{2}{\beta} + \corr{1}{2}{\beta} \right).
\end{align}

It is important that the reader understand that this is not an approximate description
of the self-assessment problem for binary classifiers. If you had the sample values for
all the variables on the right of the equal signs in eqs 9-13 , these equations would
evaluate to the exact values observed for the left variables. This comes about because
the sample statistics for the decisions of any number of classifiers exist in a finite
dimensional space. We can exhaustively enumerate all the sample statistics - observables 
and ground truths statistics - that would explain the observed agreements and disagreements
between the classifiers. This is to be contrasted with any scientific
model that is seeking to explain the observed errors. Since we do not know what model
would be correct, we cannot know if our theory of why the errors occurred is underestimating or 
overestimating the correct number of parameters. That model uncertainty does not exist
when we just consider sample statistics estimation and forego any further theoretical 
understanding of why the classifiers are making mistakes on a data stream.

\subsection{The two binary classifier self-asssessment problem is unsolvable}

The self-assessment problem for two binary classifiers is also unsolvable in
full generality. A naive counting argument shows why. Since the observable
frequencies must sum to one, there are really only 3 independent equations in this
case. But to obtain point solutions, we would have to solve for 7 variables. This result
is also formalized in Supplement A as an Algebraic Geometry algorithm.

The estimation is also impossible even if the two binary classifiers were
not correlated on the sample ($\corr{i}{j}{\ell} = 0$). In that case, we would
have to solve for 5 ground truth statistics, but again fall short of having only 3
independent equations. The unsolvability
is illustrated in Figure~\ref{figure:tryptich}(b) for one of the classifiers in the $n=2$
ensemble (classifier 1 of the {\tt twonorm} experiment), the extra information of
comparing with the other classifier (classifier 2 in the {\tt twonorm} experiment)
divides the $n=1$ surface into two disjoint pieces.
This unsolvability for independent binary classifiers
changes dramatically when we move on to consider the $n=3$ case.

\section{Algebraic Geometry of Self-Assessment for Three Binary Classifiers}

Three independent binary classifiers can solve the self-assessment problem in
certain circumstances. Before we discuss the solution, let us consider 
again the issue of general unsolvability using possibly sample-correlated 
classifiers. Once we move to three or more  classifiers there is an explosion 
in the number of variables needed to describe the non-independence between the classifiers. 
We can do a full accounting of the dimension of that space.
\begin{itemize}
    \item One environmental ground truth statistic, \prva.
    \item Two performance ground truth statistics for the accuracies of each classifier, \prs{i}{\alpha} and \prs{i}{\beta}.
    \item m ground truth statistics for the m-way correlation on each label. 
     For 2-way correlations on a label we would need n choose 2, and so on.
\end{itemize}
The number of variables (the dimension of the space) needed to describe
all sample statistics of n classifiers is,$2^{n+1} -1.$ This immediately
makes it obvious that the general self-assessment solution for n classifiers
is impossible since the event space of the decisions of n of them would be
$2^n$ for binary classification. The polynomial exact formulation is always
short by about half the number of variables we need to estimate.

If we were doing sequence ground truth statistics, instead of the point ground truth
statistics treated here, the dimension of the sample statistic space would be larger.
But, in addition, there would be an increase in the observable statistics and their
corresponding polynomial equations. These more complicated self-assessment polynomial
systems are an area of future research.

\subsection{The full solution for three independent binary classifiers}
In spite of the general unsolvability of the self-assessment problem for binary
classifiers, at $n=3$ a new phenomena arises. It is now possible to solve the GTI problem
for independent binary classifiers. The polynomial system for three independent classifiers is,
\begin{align}
\freqthree{\alpha}{\alpha}{\alpha} & = \prva  \prs{1}{\alpha}  \, \prs{2}{\alpha} \,  \prs{3}{\alpha} + \
  (1 - \prva) (1 - \prs{1}{\beta}) (1 - \prs{2}{\beta}) (1 - \prs{3}{\beta})\\
\freqthree{\alpha}{\alpha}{\beta} & = \prva  \prs{1}{\alpha}  \, \prs{2}{\alpha} \, (1 - \prs{3}{\alpha}) + \
  (1 - \prva) (1 - \prs{1}{\beta}) (1 - \prs{2}{\beta}) \prs{3}{\beta}\\
\freqthree{\alpha}{\beta}{\alpha} & = \prva  \prs{1}{\alpha}  \, (1 - \prs{2}{\alpha}) \,  \prs{3}{\alpha} + \
  (1 - \prva) (1 - \prs{1}{\beta}) \prs{2}{\beta} (1 - \prs{3}{\beta})\\
\freqthree{\beta}{\alpha}{\alpha} & = \prva  (1 - \prs{1}{\alpha})  \, \prs{2}{\alpha} \,  \prs{3}{\alpha} + \
  (1 - \prva) \prs{1}{\beta} (1 - \prs{2}{\beta}) (1 - \prs{3}{\beta})\\
\freqthree{\beta}{\beta}{\alpha} & = \prva  (1 - \prs{1}{\alpha})  \, (1 - \prs{2}{\alpha}) \,  \prs{3}{\alpha} + \
  (1 - \prva) \prs{1}{\beta} \, \prs{2}{\beta} \, (1 - \prs{3}{\beta})\\
\freqthree{\beta}{\alpha}{\beta} & = \prva  (1 - \prs{1}{\alpha})  \, \prs{2}{\alpha} \, (1 - \prs{3}{\alpha}) + \
  (1 - \prva) \prs{1}{\beta} (1 - \prs{2}{\beta}) \prs{3}{\beta}\\
\freqthree{\alpha}{\beta}{\beta} & = \prva  \prs{1}{\alpha}  \, (1 - \prs{2}{\alpha}) \,  (1 - \prs{3}{\alpha}) + \
  (1 - \prva) (1 - \prs{1}{\beta}) \prs{2}{\beta} \prs{3}{\beta}\\
\freqthree{\beta}{\beta}{\beta} & = \prva  (1 - \prs{1}{\alpha})  \, (1 - \prs{2}{\alpha}) \,  (1 - \prs{3}{\alpha}) + \
  (1 - \prva) \prs{1}{\beta} \, \prs{2}{\beta} \, \prs{3}{\beta}.
\end{align}
This polynomial system contains 7 unknown ground truth statistics, and there are 7 independent equations. 
Any algebraic geometry computer system can solve this problem. In Mathematica, this
polynomial problem can be solved in about 5 seconds on a standard laptop.
A full general solution for the independent case is also possible by calculating the Gr{\" o}bner basis
with elimination order for the monomials as mentioned before. 
This utilizes the Buchberger algorithm for computing a Grobner basis 
for the ideal of the polynomial system\cite{Cox}. For this system of polynomials, it returns a 
sequence of polynomials with progressively fewer variables. The bottom one, involving only the single \prva variable 
is a quadratic. The coefficients of the quadratic are very complicated functions of the f variables. One coefficient 
contains more than 1K terms. Nonetheless, when the f values are plugged into the quadratic it is 
simply a quadratic polynomial on \prva with integer ratio coefficients. Therefore there could be
be two real solutions with all values in the (0,1) range as required for the ground truth
integer ratios. We call roots that fall outside the (0,1) \emph{unphysical}. Details are given
in Supplement A.

This ambiguity may seem to contradict our assertion that
the self-assessment problem is solvable for independent binary classifiers. 
In the right engineering context it is possible to remove the ambiguity with side information.
For example, this ambiguity of the "decoding" for self-assessment is entirely analogous to the decoding
ambiguity in error-correcting codes for detecting and correcting bit flips in computers. 
There too, there is always more than one decoding solution. An error could be caused by
one bit flip or seven. What makes algebraic GTI practical is the same thing
that makes error-correcting codes practical. In the right engineering context, you may be
assured that a majority of the classifiers are better than random. Or the prevalence may
be environmentally stable such as when detecting rare items. In such a case, we would pick
the solution that had the rare object at 1\% of prevalence rather than at 99\%.

\section{Self-Assessment Experiments in Three Standard Binary Classification Benchmarks}

We tested the practicality of estimating unknown binary classifier performance by searching
the Penn Machine Learning Benchmarks (PMLB) database~\cite{Olson2017PMLB} to see if any benchmark fulfills the conditions that would
allow us to treat the classifiers as independent. We found three benchmarks 
({\tt twonorm}, {\tt ring}, and {\tt mushroom})t hat had enough data and showed low correlation 
between the features used for training. The results are summarized in Table~\ref{table:penn-ml}.
\begin{table}
  \caption{Self-assessment runs on three Penn ML Benchmarks~\cite{Olson2017PMLB}}
  \label{table:penn-ml}
  \centering
  \begin{tabular}{lllll}
    \toprule
    Benchmark & Prevalence     & C1 & C2 & C3\\
    \midrule
    {\tt twonorm} & 0.501(0.500)  & 0.887(0.884) & 0.885(0.880) & 0.863(0.864) \\
    {\tt ring} & 0.574(0.505)  & 0.673(0.709) & 0.602(0.703) & 0.610(0.708) \\
    {\tt mushroom} & 0.596(0.518)  & 0.948(0.814) & 0.877(0.798) & 0.732(0.782) \\
    \bottomrule
  \end{tabular}
\end{table}

Since the assumption of independence always creates two solutions, how could we decide
which one was correct for presentation in Table~\ref{table:penn-ml}? Consider
the {\tt twonorm} experiment with 7,400 predictions. The two solutions are shown in
Table~\ref{table:two-norm-results}. We picked root solution 2 with the following reasoning:
In an engineering context of a well-designed set of classifiers, it is highly likely that the classifiers 
perform better than random. Is it more likely that all three classifiers have
failed or that all three are okay? This side information is entirely analogous
to error-correcting codes picking a few bit flips versus many bit flips when
they decode an error-correcting code.

\begin{table}
  \caption{The two self-assessment solutions for a single {\tt twonorm} experiment}
  \label{table:two-norm-results}
  \centering
  \begin{tabular}{lllll}
    \toprule
    Root Solution & Prevalence     & C1 & C2 & C3\\
    \midrule
    1 & 0.499  & 0.117 & 0.119 & 0.133 \\
    2 & 0.501  & 0.887 & 0.885 & 0.863 \\
    \bottomrule
  \end{tabular}
\end{table}

In contrast, the solution obtained for the \texttt{mushroom} benchmark is a failure for
the independent polynomial system. For consistency and ease of presentation,
Table~\ref{table:penn-ml} shows only the accuracies for the most common label in each
benchmark. Let us call that $\phi_{i,\alpha}.$ But the accuracies for the $\beta$ label
returned a non-physical answer for the 1st classifier as detailed in Table~\ref{table:mushroom-results}.

\begin{table}
  \caption{The {\tt mushroom} benchmark failure on the $\beta$ label accuracies, $\phi_{i,\beta}.$}
  \label{table:mushroom-results}
  \centering
  \begin{tabular}{lll}
    \toprule
    C1 & C2 & C3\\
    \midrule
    1.001(0.916) & 0.848(0.939) & 0.868(0.742)\\
    \bottomrule
  \end{tabular}
\end{table}

The failure of mushroom should be noted as an actual strength of the algebraic approach. In
actuality, it returned non-sensical solutions. This is a clear alarm that the assumption of
classifier independence is violated in the {\tt mushroom} experiment, which
is detailed further in Supplement B.

\section{Self-Assessment Experiments for Online ID Accuracy Systems}

We now discuss a seemingly unrelated task which can, in fact, be mapped to a self-assessment
problem that is identical to the binary classifier polynomial system. We consider the problem
of measuring the accuracy of 4 different ID systems used in online advertising. This experiment
is important because, in fact, we do not have any knowledge of the true identities of the users
who were tagged with the noisy online anonymous IDs. So how could one be assured that
a solution to the independent classifier polynomial system is actually correct? The answer
is very similar to the approach of error-correcting codes.

In this experiment we have 4 noisy ID systems. Therefore, we can create 4 triplets that can then
be self-assessed with the 3 independent classifier polynomial estimator. Our claim is that
the self-consistency between the solutions obtained is that assurance. This can actually
be stated as an algebraic geometry conjecture. The
resolution of the theorem remains an open problem (see Supplement C).

Online ID systems are not binary classifiers. They must produce an arbitrarily number of IDs
that could vary from context to context. However, we can take the stream of IDs produced by
the ID systems and map it into a binary stream by just asking from each ID system if it has
observed the ID before in the data stream. This creates a series of binary labels
(\emph{new}, \emph{old}) that maps the problem directly into the binary classifier
polynomial system.

The results for the four triplet self-assessment solutions are shown in Table~\ref{table:id-systems}.
The ambiguity of the two solutions makes an automatic solution impossible. The engineering context
in this case was assumed to be that the majority of ID systems should be better than random on
identifying new users. The self-consistency between all possible triplet solutions is remarkable.
Since the prevalence of new users is precisely the unique count of entities under the ID systems,
this algorithm is actually a unique count estimator on data streams where we only have noisy
{\tt Identity} functions. 

In addition, note that it can be implemented with a tiny memory
footprint in comparison to algorithms like HyperLogLog. This is, therefore, an example where
noise actually decreases the memory footprint of an algorithm. We discuss later on the implications
of this observation for an alternative architecture of artificial intelligence. Details of this
experiment are discussed in Supplement C.

\begin{table}
  \caption{Comparing trio solutions for 4 ID systems}
  \label{table:id-systems}
  \centering
  \begin{tabular}{llllll}
    \toprule
    Triplet & Unique Users \% & $ID_1$ IP & $ID_2$ Swoop & $ID_3$ SR & $ID_4$ Vendor\\
    \midrule
    (1,2,3) & 0.797443 & 0.985041 & 0.999959 & 0.607004 & N/A \\
    (1,2,4) & 0.799948 & 0.982922 & 0.999951 & N/A & 0.995661 \\
    (1,3,4) & 0.795855 & 0.985627  & N/A & 0.607688 & 0.996777 \\
    (2,3,4) & 0.799688 & N/A  & 0.999984 & 0.605286 & 0.995952 \\
    \bottomrule
  \end{tabular}
\end{table}
\section{Conclusions}

A non-parametric error estimator of errors made by binary classifiers becomes
possible when the ensemble has three or more members. While a general solution
to the self-assessment problem for them is impossible, one can engineer
classifiers that are close enough to independence to make this method practical.
We exhibited two Penn-ML benchmarks where the independent
errors polynomial system worked, and one where it failed. The failure is further indication of the utility
of this non-parametric approach since it signals to the user that the independent assumption
is incorrect and polynomial systems that include correlation parameters need to
be used. Which contrasts with Maximum Likelihood methods that will always
return a best fit solution.

In addition, we discussed an experiment in an industrial setting where no ground
truth is available for the correct answers but the self-consistency of all possible
trio solutions from 4 binarized outputs strongly suggests the independence assumption
was correct. An algebraic geometry conjecture that will settle the question of whether
the self-consistency can only be explained by independent errors on the sample is
discussed in Supplement C.

\section*{Broader Impact}

Error is at the heart of Science. If you understand errors, you understand Science.
Tools like algebraic GTI can complement classifiers and regressors. We expect that
new examples will be developed for recommender systems, ranked lists, etc. Many
other polynomial systems remain to be discovered for the simple reason that while
there is one ground truth, there are many statistics of ground truth. We close
with some questions about the possible significance of non-parametric estimators
for Neuroscience.

Is there an algorithm that can measure IQ without having any Psychology or World
representation inside of it? We do not know. But if the representation problem
for intelligence gets one into circular arguments, why not question the necessity
of an assessor that has any intelligence? In this paper we have discussed many
algorithms that are able to monitor with the barest of theoretical assumptions.
The regression self-assessment solution by \cite{CorradaSchultz2008} could be
used to monitor the signals from other cells. If the brain is algorithmic, did
it miss this whole class of algorithms?

The work here and in papers such as \citet{CorradaSchultz2008} have broad 
applications for the safety engineering of AI systems. They allow a machine
to self-assess its own performance. As we have mentioned, while there is no
general solution to the self-assessment problems discussed here, in the right
engineering context these solutions are practical and useful.

\medskip

\small

\bibliography{neurips_2020}

\end{document}